\def\year{2021}\relax
\documentclass[letterpaper]{article} 
\usepackage{aaai21}  
\usepackage{times}  
\usepackage{helvet} 
\usepackage{courier}  
\usepackage[hyphens]{url}  
\usepackage{graphicx} 
\usepackage{natbib}  
\usepackage{caption} 
\usepackage{xcolor}
\usepackage{multirow}
\usepackage{tabularx}
\usepackage{booktabs}
\usepackage{makecell}
\usepackage{xcolor}
\usepackage[linesnumbered,ruled,vlined]{algorithm2e}

\usepackage{lipsum}

\newcommand\blfootnote[1]{%
  \begingroup
  \renewcommand\thefootnote{}\footnote{#1}%
  \addtocounter{footnote}{-1}%
  \endgroup
}

\title{A Hybrid Task-Oriented Dialog System with Domain and Task Adaptive Pretraining}

\author{
    Boliang Zhang$^{*}$\blfootnote{Boliang Zhang and Ying Lyu contributed equally to this work.}, Ying Lyu$^{*}$, Ning Ding, Tianhao Shen, Zhaoyang Jia, Kun Han, Kevin Knight
    \\
}
\affiliations{
    DiDi AI Labs\\

    \{boliangzhang, yinglu, yaeldingning, shentianhao\_i, jiazhaoyang, kunhan, kevinknight\}@didiglobal.com
}

\newcommand{\etc}{\emph{etc}}
\newcommand{\ie}{i.e.}
\newcommand{\eg}{e.g.}

\begin{document}
\maketitle

\begin{abstract}

This paper describes our submission for the End-to-end Multi-domain Task Completion Dialog shared task at the 9th Dialog System Technology Challenge (DSTC-9). Participants in the shared task build an end-to-end task completion dialog system which is evaluated by human evaluation and a user simulator based automatic evaluation. 
Different from traditional pipelined approaches where modules are optimized individually and suffer from cascading failure, we propose an end-to-end dialog system that 
1) uses Generative Pretraining 2 (GPT-2) as the backbone to jointly solve Natural Language Understanding, Dialog State Tracking, and Natural Language Generation tasks,
2) adopts Domain and Task Adaptive Pretraining to tailor GPT-2 to the dialog domain before finetuning,
3) utilizes heuristic pre/post-processing rules that greatly simplify the prediction tasks and improve generalizability, and
4) equips a fault tolerance module to correct errors and inappropriate responses.
Our proposed method significantly outperforms baselines and ties for first place in the official evaluation.
We make our source code publicly available.

\end{abstract}
\section{Introduction}


Task-oriented dialog systems aim to communicate with users through natural language to accomplish
a wide range of tasks, such as restaurant booking, weather querying, \etc. 
With the rising trend of artificial intelligence, many devices are incorporated with virtual assistants, such as Alexa, Siri, and Cortana. Task-oriented dialog systems have attracted attention from both academia and industry as a key component in virtual assistants~\citep{chen2018deep,gao2018neural}.

Real-world dialogue systems usually need to deal with complex tasks containing multiple goals and spanning over multiple domains, which poses great challenges to the existing task-oriented dialog systems.  
Traditionally a task-oriented dialog system uses a pipeline architecture that consists of the following modules: Natural Language Understanding (NLU), Dialog Manager (DM) that tracks dialog states and predicts actions, and Natural Language Generation (NLG)~\citep{williams2014dialog,bocklisch2017rasa,gao2019dialog,zhang2019find}. These modules are usually isolated and optimized individually. Therefore, errors can propagate from module to module and hurt the overall performance~\citep{ham-etal-2020-end,gao2018neural}. Further, such pipeline-based solutions usually deal with simple tasks within a single domain, requiring rich domain knowledge and expert experiences. Hence it is prohibitively expensive to build dialog systems at scale for complex tasks with multiple domains.

Fully data-driven dialog systems have been extensively studied recently due to the success of deep learning. They jointly learn to understand the user's language, inquire databases, and compose responses.
These end-to-end dialog systems do not rely on the traditional components and have shown great potentials.
\citet{wen2017network,yang2017end,ham2020end} demonstrate that end-to-end systems outperform the traditional pipeline approaches in task-oriented dialog scenarios. 
\citet{zhang2019task,peng2020soloist} focus on the benchmarks of the MultiWoz dataset and achieve top performance.

In this paper, we introduce our submission for the Multi-domain Task-oriented Dialog Challenge at Dialog System Technology Challenge 9 (DSTC9,~\citet{DSTC9}).

Participants in the shared task build end-to-end dialog systems that can assist human to fulfil single or multiple tasks, such as making a restaurant reservation, booking a hotel, etc. There are two types of evaluations: 
1) human evaluation where the organizer recruits Amazon Mechanical Turkers (MTurkers) to chat with the system and assess whether tasks are accomplished, and 2) automatic evaluation where the organizer provides a user simulator that scores each submission based on its conversation with the system. Human evaluation result is the only metric used for final ranking. 
Both MTurkers and the user simulator are provided with a clear and pre-defined user goal prior to the conversation. They chat with the system by following the user goal. 
To support this, ConvLab-2~\citep{zhu2020convlab} was released to serve as the platform for dialog system development and evaluation, providing a user simulator and evaluator so that participants can effectively run offline experiments and evaluations. 
The task provides the MultiWoz 2.1~\citep{eric2019multiwoz} dataset for the system development. 
In addition, any external datasets and resources are allowed in the shared task.

In this shared task, we adopt the idea of the end-to-end dialog system and propose several novel ideas to improve its performance in the real world scenario. There are five key components in our system:
\begin{itemize}
    
    \item \textbf{Domain/Task Adaptive Pretraining based on GPT-2} Following~\citet{peng2020soloist}, we build our dialog model initialized with GPT-2 to inherit its capability of producing human-like responses and leverage external topically related datasets (\eg, Schema-guided Dialog~\citep{rastogi2019towards}, Taskmaster~\citep{byrne2019taskmaster}) for pretraining. 
    However, unlike the conventional pretraining, 
    we apply the domain/task adaptive pretraining~\citep{gururangan2020don} on the external topically related datasets to tailor GPT-2 from raw web-texts to the dialog domain before finetuning with MultiWoz.  
    To the best of our knowledge, this is the first time to apply the this pretraining paradigm to the task-oriented dialog system.  
    This pretraining brings us non-trivial gains in the automatic evaluation.
    
    \item \textbf{Multi-Task Finetuning} As suggested by \citet{peng2020soloist}, we flatten the dialog history, belief states, database query results, and the response into one string, and then finetune the pretrained model with the MultiWoz dataset optimizing a combination of three objective: 1) to generate belief states conditioned on the dialog history; 2) to generate response conditioned on the belief states and dialog history; and 3) to distinguish gold samples from a distractor with fake samples. 
    In this way, we format the data generation process of dialog system (NLU, DST, POL, NLG) as a single neural model where the full sequence can be learned in a auto-regressive manner. 
    
    \item \textbf{Domain-Aware Data Pre/Post-processing} 
    Although the backbone of the proposed method is an end-to-end model, it is important to apply proper data pre-processing and post-processing because the input and the output of the model are not free-form natural language. 
    As the MultiWoz dataset is human-human conversation based,  we follow~\citet{wen2017network,zhang2019task} to create rules to clean up and delexicalize agent utterances into templates, such as 
    
    ``\texttt{\scriptsize it 's a hotel . there are 5 guesthouses in the area . do you prefer cheap or moderate for the price range ?}" 
    
    is transformed to: 
    
    ``\texttt{\scriptsize it is a [value\_type] . there are 5 [value\_type] in the area . do you prefer [value\_pricerange] or [value\_pricerange] for the price range ?}".
    
    It simplifies the training data and makes the model to only predict the templates.  
    To encourage knowledge sharing through semantically similar slots in different domains, we apply domain-adaptive delexicalization~\citep{zhang2019task} and compute the domain at each dialog turn to replace the placeholders with slots in the right domain. 
    %
    %

    \item \textbf{Fault Tolerance}
    We adjust GPT-2's decoder configuration, \eg increasing the beam size, to generate additional responses when errors or inappropriate responses occur.
    We observe significant gains in both automatic and human evaluations during our system development.
    
    \item \textbf{``User Interface"} 
    It is a special rule-based post-processing module that polishes agent utterances and make the conversation smooth. The modifications are only visible to the user.
    This improves the system performance a lot in the human evaluation.
\end{itemize}

Our submission significantly outperforms baseline methods and ties for first place in the official evaluation. We make our source code publicly available.\footnote{\url{https://github.com/boliangz/dstc9}}





\section{Related Work}

Building end-to-end trainable neural networks is becoming a new research trend for task-orientated dialog systems~\citep{wen2017network, lei2018sequicity, mehri2019structured}. The first efforts (\eg, \citep{mehri2019structured}) are the fusion methods, which attempted to integrate pretrained dialog modules (\ie, NLU, DM, NLG) into a neural dialog model. 
Sequicity~\citep{lei2018sequicity} is the first seq2seq architecture that integrates track dialogue believes in end-to-end task-oriented dialog. 
Though these methods have achieved promising results, they were usually designed for a specific domain, rendering difficulties in generalizing to multi-domains, \eg, the recently proposed multi-domain dataset MultiWoz~\citep{eric2019multiwoz}. 
Subsequently, there are several models are proposed to handle the multi-domain response generation task~\citep{zhao2019rethinking, chen2019semantically, qin-etal-2020-dynamic}. 
To prevent dialog acts growing combinatorially with the number of domains, ~\citet{chen2019semantically} built a multi-layer hierarchical graph to represent dialog acts to generate responses using BERT-based dialog policy. 
%
\citet{qin-etal-2020-dynamic} leveraged domain-shared features across domains and proposed a shared-private network, Dynamic Fusion Network, to learn shared and specific knowledge, explicitly capturing the correlation between domains. 
However, these works need a significant number of in-domain training examples to achieve good performance. In our system, we aim to generalize to multiple new domains with a few labelled examples via pretraining.

To the best of our knowledge, the most related works to ours are~\citet{ham-etal-2020-end} and~\citet{peng2020soloist}. 
\citet{ham-etal-2020-end}, DSTC8 Track 1 Winner,\footnote{https://convlab.github.io/about.html} is the first attempt to leverage GPT-2 to finetune on the new task-oriented dialog task. Instead of finetuning GPT-2 directly on the target domain, both \citet{peng2020soloist} and our model finetuned GPT-2 further with a large-scale external task-oriented dialog data to tailor GPT-2 to the dialog domain and then finetuned it on the target new domain. 
%
%
%
However, \citet{peng2020soloist} used the same multi-task objectives during the two stages of external dialog data pretraining and target domain data finetuning. In our system, to endow the model with task-oriented language generation ability, we apply Domain/Task adaptive pretraining, \ie, optimizing the GPT-2 language model objective on the external domain dialog data and the target MultiWoz dataset, respectively. 

\section{Method}

\begin{figure*}[h]
    \centering
    \includegraphics[width=17cm]{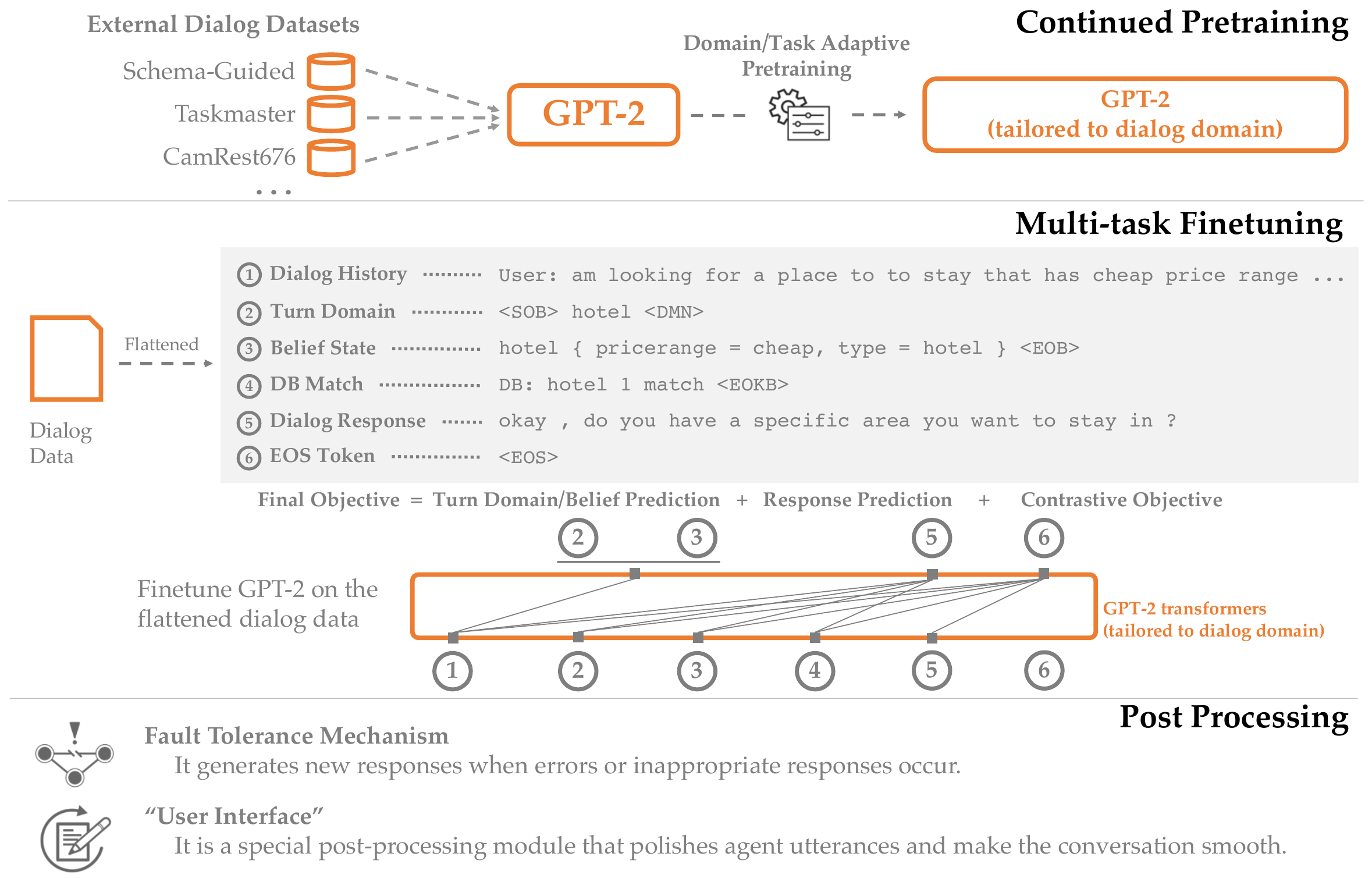}
    \caption{An overview of our system. Continued pre-training tailors GPT-2 to dialog related domain. Multi-task fine-tuning trains GPT-2 to predict turn domain, belief state and dialog response under the MultiWoz setting. Two post processing modules revise the system predictions and make them more human-readable}
    \label{fig:overview}
\end{figure*}

Figure~\ref{fig:overview} shows an overview of our system. We first pre-train the GPT-2 model on external dialog related datasets (~\ref{tab:data_stats}), simply using the language modeling objective. Then we fine-tune the dialog domain tailored GPT-2 on the MultiWoz dataset with three distinct objectives. At last, we apply a fault tolerance mechanism and a special ``user interface" to polish the predicted responses.

\subsection{The End-to-end Model}

The end-to-end model in our system consists of two parts: 1) Domain/Task adaptive pretraining that continues pretraining GPT-2 on external dialog related datasets to tailor GPT-2 to the dialog domain, and 2) finetuning GPT-2 on the MultiWoz dataset using three task specific objectives that are carefully designed to make the model learn to accurately predict belief state and response.

\textit{Domain/Task Adaptive Pretraining}

\citet{gururangan2020don} propose the idea of ``Don't Stop Pretraining" and show that it is still helpful to tailor a pretrained model to the domain of a target task. For Domain Adaptive Pretraining (DAPT), we collect publicly available dialog related datasets, such as Taskmaster and Schema-guided Dialog dataset, and pretrain GPT-2 on raw utterances using the original GPT-2 language model objective. For Task Adaptive Pretraining (TAPT), we pretrain GPT-2 only on raw utterances of the MultiWoz dataset. We show data statistics in Table~\ref{tab:data_stats}.

\textit{Multi-task Fine-tuning} 

Many recent work attempt to use end-to-end neural models, such as sequence-to-sequence and GPT-2, to solve task-oriented dialog problems and achieve remarkable results ~\citep{wen2016network,budzianowski-etal-2018-multiwoz,zhang2019task,peng2020soloist,ham-etal-2020-end}.
In our work, we follow the finetuning strategy of~\citet{peng2020soloist} where we concatenate the dialog history and annotations and flatten them into a string, and then use a combination of three objectives to fine-tune GPT-2.

Figure~\ref{fig:overview} shows an overview of the system. As shown in the ``Multi-task Fine-tuning" section of Figure~\ref{fig:overview}, we flatten pre-processed dialog data into a string of six \textbf{c}omponents:
\begin{itemize}
    \item [$\textbf{c}_1:$] \underline{Dialog History} We concatenate history utterances (plus the user utterance of the current turn) and add ``User:" and ``System:" to show the role of the utterance.
    \item [$\textbf{c}_2:$] \underline{Turn Domain} The domain of the current turn. ``\texttt{\small <SOB>}" (start of the belief state) and ``\texttt{\small <DMN>}"(domain) are special delimiters.
    \item [$\textbf{c}_3:$] \underline{Belief State} The belief state of the current turn is annotated by human. It ends with ``\texttt{\small <EOB>}"(end of the belief state)
    \item [$\textbf{c}_4:$] \underline{DB Match} We compute the number of entities that match the requirements in the belief state. It ends with ``\texttt{\small <EOKB>}"(end of the KB)
    \item [$\textbf{c}_5:$] \underline{Dialog Response} Delexicalized dialog response.
    \item [$\textbf{c}_6:$] \underline{EOS Token} ``\texttt{\small <EOS>}"(end of the string) is used to compute the contrastive objective, which is elaborated below.
\end{itemize}

During training, we adopt the multi-task fine-tuning strategy of~\citet{peng2020soloist}: 
\begin{itemize}
    \item \textbf{Belief Prediction}: We use $\textbf{c}_1$ (dialog history) to predict $\textbf{c}_2$ (turn domain) and $\textbf{c}_3$ (belief state), and define the objective as:

    \begin{displaymath}
    \setlength{\abovedisplayskip}{0pt}
    \setlength{\belowdisplayskip}{0pt}
    \resizebox{.4 \textwidth}{!}{
    $\mathcal{L}_B=\log p(\textbf{c}_2,\textbf{c}_3|\textbf{c}_1)=\mathlarger{\sum}\limits_{t=1}^{T_{\textbf{c}_{2,3}}}\log p_{\theta}(c_t | c_{<t}, \textbf{c}_1)$
    }
    \end{displaymath}
    , where $c_{<t}$ indicates all tokens before $t$.

    \item \textbf{Response Prediction}: We use $\textbf{c}_3$ (Belief State) to query the database and create $\textbf{c}_4$ (DB Match). Then we predict $\textbf{c}_5$ (Dialog Response) which is conditioned on $\textbf{c}_1$, $\textbf{c}_2$, $\textbf{c}_3$, and $\textbf{c}_4$. The response prediction objective is defined as:
    \begin{displaymath}
    \setlength{\abovedisplayskip}{0pt}
    \setlength{\belowdisplayskip}{0pt}
    \resizebox{.4 \textwidth}{!}{
    $\mathcal{L}_R=\log p(\textbf{c}_5|\textbf{c}_{1-4})=\mathlarger{\sum}\limits_{t=1}^{T_{\textbf{c}_5}}\log p_{\theta}(c_t | c_{<t}, \textbf{c}_{1-4})$
    }
    \end{displaymath}
    , where $c_{<t}$ indicates all tokens before $t$.
    
    \item \textbf{Contrastive Ojbective}: It is a common practice to introduce constrastive samples for training machine learning models~\citep{peng2020soloist,ham-etal-2020-end}. We create the negative sample by modifying the positive sample, through one of the following three ways: 1) replace $\textbf{c}_2$ (belief state) with another random $\textbf{c}_2$, 2) replace $\textbf{c}_5$ (dialog response) with another random $\textbf{c}_5$, or 3) replace both $\textbf{c}_2$ and $\textbf{c}_5$ with another $\textbf{c}_2$ and $\textbf{c}_5$. We apply a binary classifier on ``\texttt{\small <EOS>}" token of the positive and negative samples. The objective function is:
    \begin{displaymath}
    \setlength{\abovedisplayskip}{1pt}
    \setlength{\belowdisplayskip}{1pt}
    \resizebox{.45 \textwidth}{!}{
    $\mathcal{L}_C=y \log (p_{\theta}(\textbf{positive})) + (1-y)\log(1-p_{\theta}(\textbf{negative}))$
    }
    \end{displaymath}
\end{itemize}

Thus, the full fine-tuning objective is:
\begin{displaymath}
\setlength{\abovedisplayskip}{1pt}
\setlength{\belowdisplayskip}{1pt}
\resizebox{.18 \textwidth}{!}{
$\mathcal{L}=\mathcal{L}_B+\mathcal{L}_R+\mathcal{L}_C$.
}
\end{displaymath}

During inference, there are two stages of predictions: 1) given $\textbf{c}_1$ (dialog history), the model predicts $\textbf{c}_2$ (turn domain) and $\textbf{c}_3$ (belief state), and queries the database to generate $\textbf{c}_4$ (DB match), and 2) the model predicts $\textbf{c}_5$ (dialog response) based on $\textbf{c}_{1-4}$.


\subsection{Pre/Post-processing}


We utilize heuristic pre/post-processing rules that largely simplify the prediction tasks and improve generalizability.

\begin{itemize}
\item \textbf{Domain-Adaptive Delexicalization} 
In order to address the problem of massive entity number in the system response, we follow the delexicalization pipeline suggested by~\citet{wen2017network} to generate delexicalized sentences with placeholders during pre-processing, and then apply the post-processing module to the predicted sentence by replacing the placeholders with the corresponding DB record. 
%
%
However, for the MultiWoz dataset, there are many slots which exist in multiple domains. For example, slots \textit{name}, \textit{type} and \textit{address} exist in both domains \textit{restaurant} and \textit{hotel}. It results in great burden for the system to generate the placeholder tokens if delexicalizing the same slot into different placeholders, \eg, $[restaurant\_name]$ and $[hotel\_name]$. To address this problem, we apply the domain-adaptive delexicalization~\citep{zhang2019task} that uses an identical placeholder $[value\_name]$ to encourage knowledge sharing through semantically similar slots in different domains. Since our model could predict domain for each turn, there is no ambiguity in the post-processing stage for the placeholder replacement.


\vspace{-2mm}

\item \textbf{Turn Domain Computing} Turn domain is referred as the domain involved in the current dialog turn. It is critical for a model to be able to predict an exact turn domain to facilitate post-processing, \eg, a domain-adaptive placeholder could be replaced with the slot in the right domain. To this end, we need compute the turn domain from the MultiWoz dataset to feed to the model during training. 
There is no turn domain label in the dataset. Although it may involve multiple domains in one dialog, domains in a dialog are usually not changed back and forth. Based on this feature, we could compute turn domains by tracking the changes of the labeled belief states among the dialog turns. 
Specifically, for each domain with non empty constraint in the belief state of the current dialog turn, if the domain is a new domain, \ie, is not mentioned in the dialog history yet, or its corresponding constraint has been updated in the current turn, then the domain is appended to the turn domain. 
The computed turn domain is empty, then inherit from the turn domain at the previous turn which is initialized as the domain \texttt{general}. 
%
As at each dialog turn in MultiWoz it usually involves only one domain, the final computed turn domain includes one domain.


\vspace{-2mm}
\item \textbf{Data Cleaning and Normalization} We apply the following steps to clean the dataset during pre-processing. 
First, all the dialog utterances are set to be lower case for easier manipulations. 
Additionally, dialogs are processed as a set of dialog turns, each of which is in the form of one user utterance and one system response. Consecutive user utterances between system responses are merged into one. Recall that model training samples are composed of dialog history, belief state, database state and response. To get more dialog contexts, we maximize the number of dialog turns in the dialog history to keep each sample length within 512. 
Moreover, different slot names/values with the same semantics are normalized to the same slot/value. For example, \texttt{departure} and \texttt{pickup\_location} are normalized to be \texttt{departure}.    
%
We apply this data cleaning and normalization process to MultiWoz and all of the external datasets used for pretraining.


\end{itemize}




\subsection{Fault Tolerance Mechanism}
When the system fails or produces inappropriate responses, we increase the beam size from the default which is 1 of the GPT-2 decoder to produce a different response to correct the error. 
We list a few representative errors and the corrected results after using this mechanism, sampled from our system outputs:
\begin{itemize}
    \item Some predicted belief states are malformed. The comma in the following predicted belief state breaks the dictionary data structure and throws an exception.
    
    \texttt{\scriptsize \textbf{Dialog Context:}}
    
    \texttt{\scriptsize \underline{User:} What 's the entrance fee for abbey pool, and astroturf pitch ?}

    \texttt{\scriptsize \textbf{Before:}}
    
    \texttt{\scriptsize \underline{Predicted belief:} \{ name : abbey pool \textcolor{red}{\textbf{\underline{,}}} and astroturf pitch , area : north \}}
    
    \texttt{\scriptsize \textbf{After:}}
    
    \texttt{\scriptsize \underline{Predicted belief:} \{ name : abbey pool and astroturf pitch , area : north \}}

    \item The system may predict bad templates where the placeholders cannot be replaced by any of the slots in the belief states. In the following example, placeholder ``\texttt{\scriptsize [value\_day]}" cannot be replaced as slot type ``\texttt{\scriptsize day}" is not in the belief states. Therefore, the system uses the second ranked candidate of the decoder instead and makes a correct response.

    \texttt{\scriptsize \textbf{Dialog Context:}}

    \texttt{\scriptsize \underline{User:} Can you book me a table for 4 people? Can you try 13:00 ?}
    
    \texttt{\scriptsize \underline{Predicted belief:} restaurant \{ time = 13:00, people = 4 \}}
    
    \texttt{\scriptsize \textbf{Before:}}

    \texttt{\scriptsize \underline{Predicted template:} i have made your reservation for \textcolor{red}{\textbf{[value\_day]}} at 13:00 . your reference number is : [value\_ref] .}
    
    \texttt{\scriptsize \textbf{After:}}
    
    \texttt{\scriptsize \underline{Predicted template:} what type of food do you like ?}
    
    \item Sometimes the system constantly repeat its responses and lead to bad user experience. It may also cause the dialog fall off the track. The fault tolerance mechanism is triggered when the system generates responses similar to its last one (fuzzy match is used to measure the similarity).
    
    \texttt{\scriptsize \textbf{Dialog Context:}}
        
    \texttt{\scriptsize \underline{User:} i want a place to stay.}

    \texttt{\scriptsize \underline{Sys:} okay , what area of town would you like to stay in ?}

    \texttt{\scriptsize \underline{User:} i don't have preference.}

    \texttt{\scriptsize \textbf{Before:}}
    
    \texttt{\scriptsize \underline{Sys:} okay , i can help you with that . what area of town would you like to stay in ?}

    \texttt{\scriptsize \textbf{After:}}
    
    \texttt{\scriptsize \underline{Sys:} i have 33 guest houses in the east . do you have a price range ?}
\end{itemize}


\subsection{``User Interface"}

In the human evaluation, we notice that some errors are due to the missing information in responses.
``User Interface" is a special post-processing module to let us manually revise the predicted responses to include more information to improve user experience. Note that to avoid changing the data distribution, the original utterances are used as the dialog history for the next turn prediction, rather than the ``User Interface" revised responses. 
We show some examples below:

\texttt{\scriptsize \textbf{Dialog Context}}

\texttt{\scriptsize \underline{User:} i 'll be leaving from peterborough and going to cambridge on tuesday , sometime after 15:15 . could you book that for me for 2 people ?}

\texttt{\scriptsize \textbf{Before:}}

\texttt{\scriptsize \underline{Sys:} sure thing . i 've booked two tickets on tr5933 . the total fee is 33 gbp and your reference number is tl4r46ys .}

\texttt{\scriptsize \textbf{After:}}

\texttt{\scriptsize \underline{Sys:} booking was successful . departure: peterborough, destination: cambridge, leave time: 15:19, people: 2, price: 33 pounds, reference number: tl4r46ys . is there anything else i can help with ?}

In this example, the system correctly books two train tickets based on the user's requirement, while it does not provide the full information of the train to the user and cannot convince the user that it has booked the right train. In the human evaluation error analysis, we notice that some users may further ask questions to confirm the booking which may cause the system to fall off the track. This type of error occurs in booking in the train, hotel, and restaurant domains. Therefore, we correct the system so that it provides more booking information to the user to solve the problem.

\texttt{\scriptsize \textbf{Dialog Context:}}

\texttt{\scriptsize \underline{User:} i would like to find someplace to visit .}

\texttt{\scriptsize \textbf{Before:}}

\texttt{\scriptsize \underline{Sys:} what area of the town do you prefer ?}

\texttt{\scriptsize \textbf{After:}}

\texttt{\scriptsize \underline{Sys:} what area of the town do you prefer ? for example north, south, or center ?}

In the above example, the user is supposed to provide the system the area of the place, which should be one of the pre-defined values: north, south, west, east and center. However, sometimes the user goal does not include the area information and the users do not know the pre-defined values, so they may reply with some requirements that the system cannot handle, such as ``\texttt{\scriptsize is there anything by the river?}" or ``\texttt{\scriptsize i prefer anything close to the mountain.}". For such cases, we let the system provide a few options to let the user choose, or at least inform the user what types of choices they may have. We apply this rule to many scenarios, for example, attraction type and area, hotel price range, restaurant food type, etc. This rule is triggered when the system response matches some certain patterns, such as ``the utterance starts with which/what/where and ends with question mark".

This user interface is designed to solve problems in the human evaluation as generally as possible. It does not affect the automatic evaluation but is very helpful in the real world scenario.

\section{Experiments and Results}

\subsection{Datasets}
\begin{table}[h!t]
    \centering
    \footnotesize
    \begin{tabular}{p{2cm}p{1cm}p{1cm}c}
        \hline
        \textbf{Name} &\textbf{\#Dialog} & \textbf{\#Utterance}& \textbf{Domains} \\
        \hline
        \multirow{3}{*}{MultiWoz 2.1}        & \multirow{3}{*}{10,421} & \multirow{3}{*}{142,840}  &Restaurant, Police,\\
                                             & & & Hotel, Train, Taxi,\\
                                             & & & Attraction, Hospital\\
        \hline
        \multirow{2}{*}{Schema}              & \multirow{2}{*}{6969} & \multirow{2}{*}{50,192}    &Restaurants, Hotels,\\
                                             & & & Trains, Travel\\
        \hline
        CamRest          & 676    & 5488    &Restaurant\\
        \hline
        \multirow{3}{*}{Taskmaster2020}      & \multirow{3}{*}{5873} & \multirow{3}{*}{98,662}       &Restaurant, Food,\\
                                             & & & Hungry, Dessert,\\
                                             & & & Lunch, Dinner, Hotel\\
        \hline
        Taskmaster2019 & 4349   & 89,076       &Uber, Restaurant\\
        \hline
        MSR-E2E             & 6969 & 50,192  &Restaurant, Taxi\\
        \hline
    \end{tabular}
    \vspace{-2mm}
    \caption{ Statistics of dialog corpora used in our system}
    \label{tab:data_stats}
\end{table}

\noindent\textbf{Dataset for Finetuning} The dataset model finetuning is MultiWoz 2.1~\citep{eric2019multiwoz}, which a large-scale human-human multi-domain task-oriented dialog dataset. It contains 8421/1000/1000 for training/validation/testing, respectively. 
As this challenge does not utilize the testing data for evaluation, we append it into the training data and use the two datasets for the model finetuning. 


\noindent\textbf{Dataset for Pretraining} Unlike the model finetuning samples that are composed of dialog history, belief state, database state and response, the datasets for pretraining include user or system utterances only, excluding belief state and database state information. The rationale for this is 1) the optimization objective for pretraining is based on language model; 2) external dialog datasets have good quality dialog utterances but have poor quality labels for belief states or database information. 
%
%
In our experiments, the Task-Adaptive Pretraining (TAPT) dataset is based on MultiWoz. 
%
%
For Domain-Adaptive Pretraining (DAPT), we use the external dialog data, \ie, Schema~\citep{rastogi2019towards}, CamRest~\citep{wen2016conditional}, Taskmaster~\footnote{https://github.com/google-research-datasets/Taskmaster} and MSR-E2E~\footnote{https://github.com/xiul-msr/e2e\_dialog\_challenge}, as shown in Table~\ref{tab:data_stats}. As suggested in~\citet{gururangan2020don}, the DAPT corpus from the similar domain as the target domain could improve the performance whereas the irrelevant ones may even worsen the performance. Therefore, we extract the similar domains from the external dialog data as the ones in MultiWoz. For example, for Schema, there are 17 domains in total in the raw dataset, but we use the four domains Restaurants, Hotels, Trains, Travel (similar to the Attraction domain in MultiWoz). 
%
Each sample of the DAPT/TAPT dataset is dialog utterance within 512.

\begin{table}  
    \centering
    \begin{tabular}{lr}
    \hline
    \textbf{Hyper-parameter} & \textbf{Value} \\
    \hline
    Max sequence length* & 512 \\
    Max response length$^{\dagger}$ & 128 \\
    Optimizer & SGD \\
    Learning rate & 0.001 \\
    Linear warm-up step & 500 \\
    Early stopping countdown$^{\ddagger}$ & 3 \\
    \hline
    \multicolumn{2}{p{0.4\textwidth}}{\small * Sequences longer than 512 are truncated from the head. Based on our calculation, 96\% of the dialogs in MultiWoz can fit in 512 tokens.} \\
    \multicolumn{2}{p{0.4\textwidth}}{\small $^{\dagger}$ User or system utterances longer than 128 tokens are truncated from the tail.} \\
    \multicolumn{2}{p{0.4\textwidth}}{\small $^{\ddagger}$ Training is terminated if the loss on development set does not decrease for three evaluations. The model converges in about three epochs.} \\
    \end{tabular}
    \vspace{-2mm}
    \caption{Hyper-parameters}
    \label{tab:hype_param}
\end{table}

\begin{table*}[h!t]
    \centering
    \begin{tabular}{ccrrrrrr}
        \hline
        \multirow{2}{*}{\textbf{Rank}} &
        \multirow{2}{*}{\textbf{Team}} & \multicolumn{3}{c}{\textbf{Success Rate}} & \multirow{2}{*}{\textbf{Language}} & \multirow{2}{*}{\textbf{Response}} & \multirow{2}{*}{\textbf{Turns}} \\
        \cmidrule(l){3-5}
        & & Avg.$^{\dagger}$ & w/ DB & w/o DB & \textbf{Understanding} & \textbf{Appropriateness} & \\
        \hline
        10 & Team 10 & 19.5 & 6.0 & 33.0 & 3.23 & 2.93 & 18.8 \\
        9 & Team 8 & 35.0 & 26.0 & 44.0 & 3.27 & 3.15 & 18.5 \\
        8 & Team 9 & 55.2 & 43.2 & 67.2 & 4.15 & 3.98 & 19.2 \\
        7 & Team 5 & 58.4 & 50.4 & 66.4 & 4.15 & 4.06 & 19.7 \\
        6 & Team 4 & 60.3 & 51.4 & 69.2 & 4.49 & 4.22 & 17.7 \\
        5 & Team 3 & 67.8 & 60.0 & 75.6 & 4.56 & 4.42 & 21.0 \\
        4 & Team 6 & 70.6 & 60.8 & 80.4 & 4.41 & 4.41 & 20.1 \\
        3 & Team 7 & 72.3 & 62.0 & 82.6 & 4.53 & 4.41 & 17.1 \\
        1 & Team 1 & 74.8 & 70.2 & 79.4 & 4.54 & 4.47 & 18.5 \\
        \hline
        \textbf{1} & \textbf{Ours} & \textbf{74.8} & \textbf{68.8} & \textbf{80.8} & \textbf{4.51} & \textbf{4.45} & \textbf{19.4} \\
        \hline
        \multicolumn{8}{p{0.7\textwidth}}{\small $^{\dagger}$ Ranking of the teams is based on the average success rate.}
    \end{tabular}
    \vspace{-2mm}
    \caption{Official results of the human evaluation. We tie for first place with Team 1. The rank is based on the average success rate. Please refer to the Human Evaluation section for details of the evaluation metric.}
    \label{tab:human_eval}
\end{table*}

\begin{table*}[h!t]
    \centering
    \begin{tabular}{ccrrrrrrr}
        \hline
        \multirow{2}{*}{\textbf{Rank}} &
        \multirow{2}{*}{\textbf{Team}} & \multirow{2}{*}{\textbf{Success}} & \multirow{2}{*}{\textbf{Book}} & \multicolumn{3}{c}{\textbf{Inform Rate}} & \multicolumn{2}{c}{\textbf{Turns}} \\ 
        \cmidrule(l){5-7}\cmidrule(l){8-9}
        & & \textbf{Rate$^{\dagger}$} & \textbf{Rate} & P & R & F & succ. & all \\
        \hline
        10 & Team 10 & 21.4 & 0.0 & 55.4 & 60.0 & 54.1 & 11.0 & 25.9 \\
        9 & Team 9 & 44.4 & 26.5 & 57.9 & 64.5 & 58.9 & 12.2 & 14.6 \\
        8 & Team 8 & 52.6 & 66.7 & 57.5 & 80.7 & 64.8 & 13.2 & 22.5 \\
        7 & Team 7 & 57.8 & 85.0 & 68.7 & 81.6 & 72.6 & 13.7 & 16.4 \\
        6 & Team 6 & 67.7 & 90.8 & 70.4 & 85.6 & 75.2 & 12.8 & 14.2 \\
        5 & Team 5 & 83.3 & 89.1 & 81.1 & 90.3 & 83.5 & 13.5 & 13.8 \\
        4 & Team 4 & 89.8 & 96.3 & 72.4 & 96.0 & 80.1 & 15.1 & 15.8 \\
        3 & Team 3 & 90.8 & 96.7 & 81.0 & 95.4 & 85.9 & 13.4 & 13.6 \\
        1 & Team 1 & 93.0 & 94.6 & 84.1 & 96.2 & 88.1 & 12.5 & 12.7 \\
        \hline
        \textbf{2} & \textbf{Ours} & \textbf{91.4} & \textbf{96.9} & \textbf{80.2} & \textbf{97.3} & \textbf{86.0} & \textbf{15.3} & \textbf{15.7} \\
        \hline
        \multicolumn{8}{p{0.5\textwidth}}{\small $^{\dagger}$ Ranking of the teams is based on the success rate.}
    \end{tabular}
    \vspace{-2mm}
    \caption{Official results of the user simulator based automatic evaluation. The rank is based on the success rate. Please refer to the Automatic Evaluation section for details of the evaluation metric.}
    \label{tab:auto_eval}
\end{table*}

\begin{table}[h!t]
    \centering
    \small
    \begin{tabular}{lrr}
        \hline
        \multirow{2}{*}{\textbf{Method}} & \textbf{Auto. Eval.}  \\
        & \textbf{Success Rate (\%)} \\
        \hline
        Vanilla GPT-2 & 88 \\
        \hspace{2mm} w/ DAPT & 91 \\
        \hspace{2mm} w/ TAPT & 86 \\
        \hspace{2mm} w/ DAPT + TAPT & 91 \\
        \hline
    \end{tabular}
    \vspace{-2mm}
    \setlength{\belowcaptionskip}{-13pt}
    \caption{Ablation study of Domain Adaptive Pretraining (\textbf{DAPT}) and Task Adaptive Pretraining (\textbf{TAPT}). We use DAPT or/and TAPT to tailor GPT-2 to dialog domain and then finetune it on MultiWoz. The table shows that DAPT improves the automatic evaluation success rate by 3\%.
    }
    \label{tab:ablation}
\end{table}

\subsection{Training Details}

We use Huggingface~\cite{wolf2019huggingface} \texttt{GPT-2-small}~\footnote{https://github.com/huggingface/transformers} in our system. It has 124M parameters, consists of 12 transformer decoder blocks, and is pre-trained on a large web-crawled dataset of various domains~\citep{radford2019language}. 
The sub-word tokenizer we use is also from Huggingface transformers module.

We use the same configurations for the Domain/Task Adaptive pre-training and finetuning on the MultiWoz dataset. 
Training details are listed in Table~\ref{tab:hype_param}.

\subsection{Human Evaluation}

Amazon Mechanical Turkers (MTurkers) are recruited to chat with the system to accomplish one or multiple tasks by following a pre-defined user goal. After the conversation, MTurkers annotate whether the tasks are successfully accomplished (Success or Fail), and rate the \textbf{Language Understanding} and  \textbf{Response Appropriateness} of the systems with scores from 1 to 5. Also, the number of \textbf{Turns} of the conversation are measured.

As MTurkers have no access to the database, when provided entity related information such as train ticket ID/price, restaurant name, etc, they cannot verify the correctness of the information. Thus, MTurkers are asked to note down all entity related information at the end of the conversation and then the shared task organizers ground this information to the database to verify their correctness. In the official evaluation, three success rate related metrics are reported:
\begin{itemize}
    \item \textbf{Success Rate w/o DB Grounding:} Annotation provided by MTurkers (Success or Fail).
    \item \textbf{Success Rate w/ DB Grounding:} The dialog is a success only if 1) MTurks mark it as ``Success", and 2) the provided entity related information can be found in the database.
    \item \textbf{Average Success:} The average of the above two scores.
\end{itemize}

The ranking of the participants is based on Average Success. Table~\ref{tab:human_eval} shows the official ranking. We tie for first place with Team 1.


\subsection{Automatic Evaluation}

To ease system development and setup an objective way to evaluate the system, ConvLab2 provides a user simulator based evaluation where the simulator talks to the system and measures the performance using several metrics. In the official evaluation, user simulator based automatic evaluation scores are reported for reference.

The user simulator is a pipelined dialog system that consists of: 1) BERTNLU that parses agent utterances into structured information, such as dialog act and slot name/value, 2) Rule-based Policy that generates user acts using pre-defined rules, and 3) Template NLG that generates natural languages utterances using 900+ templates.

Evaluation metrics include:
\begin{itemize}
    \item \textbf{Success Rate} A dialog is considered as Success only if all \textit{informable} and \textit{requestable} slots are correctly filled.
    \item \textbf{Book Rate} The average booked entities satisfy the goal constraints among all domains.
    \item \textbf{Inform Rate} Precision, recall and F-1 scores of how the user requestable slots are filled.
    \item \textbf{Turns} The number of turns of successful dialogs and all dialogs.
\end{itemize}

Table~\ref{tab:auto_eval} shows the official automatic evaluation results. We rank in second place.
Table~\ref{tab:ablation} shows an ablation study of the impact of Domain Adaptive Pretraining (\textbf{DAPT}) and Task Adaptive Pretraining (\textbf{TAPT}). We use DAPT or/and TAPT to pretrain GPT-2 on external dialog related datasets, and then finetune the dialog domain tailored GPT-2 on MultiWoz dataset. DAPT improves the system by 3\% accuracy over the baseline, while TAPT hurts the system performance. The combination of DAPT and TAPT achieves similar performance as DAPT. As we use different random seed from the official evaluation, the scores in Table~\ref{tab:ablation} are not official and not comparable with the official evaluation results.




\section{Conclusions and Future Work}

In this paper, we introduce our submission for the End-to-end Multi-domain Task Completion Dialog shared task of the DSTC9. Our method proposes to use Domain Adaptive and Task Adaptive pretraining to tailor the GPT-2 model to dialog domain. Then we finetune it on the MultiWoz dataset using three different task specific objectives. At last, we use a fault tolerance mechanism and a special user interface to polish system predictions and make them more human readable. Our system ties for first place in the competition. 
Future work can focus on optimizing database grounding which can make the information in the response more consistent to the database.
\section{Acknowledgements}
We would like to thank Arkady Arkhangodsky, Han Zhao and Jianwei Liu for their helpful comments and human evaluation. 


\bibliography{aaai21}

\end{document}